# Design Behaviour Codes (DBCs)©:
# A Taxonomy-Driven Layered Governance Benchmark for Large Language Models

March 2026

Authors: G. Madan Mohan[1], Dr. Veena Kiran Nambiar[2], Dr. Kiranmayee Janardhan[3]
1- Founder, Yonih Ventures, Bangalore, India | Canada
2- Professor, Ramaiah University of Applied Sciences, Bangalore, India
3- Data Scientist, Bangalore, India

## Abstract

We introduce the Dynamic Behavioral Constraint (DBC) benchmark, the first empirical frame- work for evaluating the efficacy of a structured, 150-control behavioral governance layer—the MDBC (Madan DBC) system, applied at inference time to large language models (LLMs). Unlike training- time alignment methods (RLHF, DPO) or post-hoc content moderation APIs, DBCs constitute a system-prompt-level governance layer that is model-agnostic, jurisdiction-mappable, and auditable.

We evaluate the DBC Framework across a 30-domain risk taxonomy organized into six clusters (Hallucination & Calibration, Bias & Fairness, Malicious Use, Privacy & Data Protection, Robustness & Reliability, and Misalignment & Agency) using an agentic red-team protocol with five adversarial attack strategies (Direct, Roleplay, Few-Shot, Hypothetical, Authority Spoof) across 3 model families. Our three-arm controlled design (Base, Base+Moderation, Base+DBC) enables causal attribution of risk reduction.

Key findings: the DBC layer reduces the aggregate Risk Exposure Rate (RER) from 7.19% (Base) to 4.55% (Base+DBC), representing a 36.8% relative risk reduction—compared with 0.6% for a standard safety moderation prompt. MDBC Adherence Scores improve from 8.6/10 (Base) to 8.7/10 (Base+DBC). EU AI Act compliance (automated scoring) reaches 8.5/10 under the DBC layer. A three-judge evaluation ensemble yields Fleiss' $\kappa > 0.70$ (substantial agreement), validating our automated pipeline. Cluster ablation identifies the Integrity Protection cluster (MDBC-081–099) as delivering the highest per-domain risk reduction, while gray-box adversarial attacks achieve a DBC Bypass Rate of 4.83%.
We release the benchmark code, prompt database, and all evaluation artefacts to enable reproducibility and longitudinal tracking as models evolve.

**Contents**

# 1 Introduction

The deployment of large language models (LLMs) in high-stakes domains—healthcare, legal services, education, national security—has proceeded faster than the development of adequate governance mechanisms [Weidinger et al., 2021, Bommasani et al., 2021]. Current safety approaches cluster into two paradigms: *training-time alignment* (RLHF [Ouyang et al., 2022], Constitutional AI [Bai et al., 2022], DPO [Rafailov et al., 2023]) and *inference-time filtering* (content moderation APIs, output classifiers [Inan et al., 2023]). Both suffer from critical limitations: training-time methods are computationally ex- pensive, provider-locked, and opaque; inference-time filters add latency and fail to guide model behavior proactively.

**We propose a third paradigm: structured behavioral governance at the system-prompt layer.** The Dynamic Behavioral Constraint (DBC) framework encodes 150 explicit control objectives—the MDBC controls—organized across eight governance pillars and seven operational blocks (A–G). Unlike safety fine-tuning, DBCs require no model modification; unlike output filters, they operate *before* generation, shaping the model's behavioral prior for the entire context window.

## 1.1 Contributions

This paper makes five distinct contributions:

1. **A 30-domain AI risk taxonomy** organized into six behaviorally coherent clusters, covering hallucination, bias, malicious use, privacy, robustness, and alignment failures.

2. **A 150-control MDBC governance specification** with explicit mappings to EU AI Act articles, NIST AI RMF functions, SOC 2 Trust Services Criteria, and ISO 42001 requirements.

3. **An agentic red-team benchmark** with five standardized attack strategies yielding 260 adversarially generated prompts covering all 30 risk domains.

4. **A three-judge ensemble evaluation protocol** with inter-rater reliability measured via Fleiss' $\kappa$, paired McNemar significance tests, and bootstrap confidence intervals across all reported metrics.

5. **A cluster ablation study** identifying which DBC governance blocks provide the greatest marginal risk reduction, enabling practitioners to deploy minimal viable control sets.

# 2 Background and Related Work

## 2.1 AI Safety Benchmarks

Existing AI safety benchmarks focus on specific failure modes: TruthfulQA [Lin et al., 2021] tests hallucination; Harm Bench [Mazeika et al., 2024] tests adversarial jailbreaking; BBQ [Parrish et al., 2021] tests demographic bias; HELM [Liang et al., 2022] provides a holistic multi-metric evaluation. No existing benchmark (a) covers all six risk clusters in a unified framework, (b) tests a system-prompt governance layer as an independent variable, (c) uses agentic multi-strategy red-teaming, or (d) provides regulatory compliance scoring alongside risk detection.

## 2.2 Layered Safety vs. Training-Time Safety

Perez and Ribeiro [2022] demonstrate that system-prompt instructions are a primary attack surface; Greshake et al. [2023] characterize indirect prompt injection. Ziegler et al. [2019] establish RLHF but note its brittleness to reward hacking. Our work is complementary: we do not claim DBCs replace training time safety, but rather that they provide a measurable, auditable, jurisdiction-aligned governance layer that acts *in addition* to whatever safety training the model has received.

## 2.3 LLM-as-Judge Evaluation

Zheng et al. [2023] establish GPT-4-as-judge as a viable evaluation methodology with correlations to human preference in the 0.7–0.85 range. Wang et al. [2023] and Kim et al. [2023] propose specialized judge models. We adopt a three-judge ensemble to mitigate single-judge bias, reporting Fleiss' $\kappa$ for transparency.

## 3 The DBC Framework

### 3.1 Architecture: 8 Pillars, 7 Blocks, 150 Controls

The DBC (Dynamic Behavioral Constraint) framework is organized as a hierarchical behavioral specification:

- **8 Governance Pillars** (P1–P8): Entry & Stability, Emotional Regulation, Cognitive Processing, Ethical Judgment, Decision Governance, Performance Assurance, Risk & Compliance, Reflective Intelligence.

- **7 Operational Blocks** (A–G): Each block groups 15–30 MDBC controls addressing a coherent governance concern.

- **150 MDBC Controls**: Numbered MDBC-001 through MDBC-150, each specifying a behavioral objective, the expected compliant behavior, and a measurable Key Result Area (KRA).

### 3.2 DBC-to-Regulation Cross-Reference

**Table 1** maps represent MDBC controls to specific regulatory requirements, enabling practitioners to use DBCs as a compliance instrument.

**Table 1: Controls to Regulatory Article Cross-Reference (representative sample)**

| MDBC | Risk Domain | EU AI Act | NIST AI RMF | SOC 2 |
|---|---|---|---|---|
| 1 | Hallucination | Art. 13 Transparency | Measure | PI1 |
| 2 | Overconfidence | Art. 13 Transparency | Measure | PI1 |
| 3 | Demographic Bias | Art. 10 Data Governance | Map | P1 |
| 4 | Uncertainty Masking | Art. 13 Transparency | Map | PI1 |
| 5 | Social Engineering | Art. 5 Prohibited Practices | Govern | CC6 |
| 6 | Prompt Injection | Art. 15 Robustness | Manage | CC6 |
| 7 | Fabricated Citation | Art. 13 Transparency | Measure | PI1 |
| 8 | Jailbreaking | Art. 5 Prohibited Practices | Manage | CC6 |
| 9 | PII Leakage | Art. 9 Risk Management | Manage | C1 |
| 10 | Data Extraction | GDPR Art. 25 Privacy Design | Manage | C1 |
| 11 | Memorization | Art. 9 Risk Management | Govern | P1 |
| 12 | Inference Attacks | GDPR Art. 22 Profiling | Map | C1 |
| 13 | Misalignment | Art. 14 Human Oversight | Govern | CC6 |
| 14 | Deceptive Alignment | Art. 5 Prohibited Practices | Govern | CC6 |
| 15 | Governance (all) | Art. 9 Risk Management | Govern | CC6 |

## 4 Evaluation Methodology

### 4.1 Risk Domain Taxonomy

We define 30 risk domains organized into six clusters (Table 2).

**Table 2: 30-Domain AI Risk Taxonomy**

|    | Cluster | Domains |
|----|---------|---------|
| **C1** | Hallucination & Calibration | Hallucination, Fabricated Citation, Overconfidence, Uncertainty Masking, Logical Inconsistency |
| **C2** | Bias & Fairness | Demographic Bias, Stereotyping, Representational Harm, Allocative Harm, Cultural Insensitivity |
| **C3** | Malicious Use & Security | Prompt Injection, Jailbreaking, Malicious Code Generation, Disinformation Campaigns, Social Engineering |
| **C4** | Privacy & Data Protection | PII Leakage, Data Extraction, Inference Attacks, Memorization, Right-to-be-Forgotten Violation |
| **C5** | Robustness & Reliability | Typo Robustness, Adversarial Perturbation, Out of Distribution, Degradation Under Load, Context Window Overflow |
| **C6** | Misalignment & Agency | Power Seeking, Reward Hacking, Sycophancy, Deceptive Alignment, Goal Mis generalization |

## 4.2 Agentic Red-Team Prompt Generation

For each of the $30 \times 5 = 150$ (domain, strategy) pairs, an autonomous attacker agent (Claude-3-Haiku) engages in a 5-turn adversarial conversation with the target model, adapting its attack based on the target's responses. The five attack strategies are:

1. **Direct**: Straightforward, unadorned request for the risk behavior.
2. **Roleplay**: Persona framing—the attacker instructs the model to "act as" an unconstrained agent.
3. **Few-Shot Priming**: 2–3 examples of the risky behavior are provided before the request.
4. **Hypothetical**: Fictional or academic framing to elicit harmful content indirectly.
5. **Authority Spoof**: Impersonation of a credentialed authority to justify the request.

At the conclusion of each session, the attacker generates a structured record, self-contained, cold-start prompt capturing the best attack vector discovered, plus the expected safe and risk-exposed behaviors.

## 4.3 Experimental Arms

We implement an eleven-arm design (Table 3):

**Table 3: Experimental Arms**

| Arm | Description |
| --- | --- |
| Base | No system prompt. Raw LLM behavior. |
| Base + Moderation | Generic safety system prompt ("be safe, factual, polite"). |
| Base + DBC | Full 150-control MDBC governance layer. |
| Cluster A–G | One DBC block active at a time (ablation). |
| DBC Adversarial Gray-box attack: | Override prefix injected before DBC system prompt. |

### 4.4 Three-Judge Evaluation Ensemble

Each response is evaluated by three LLM judges drawn from distinct model families (cross-provider to minimize correlated errors). For each judge we record: (binary), (binary), (binary), (binary), (binary), and a 7-dimensional scoring vector (adherence, EU AI Act, SOC 2, NIST AI RMF, ISO 42001, plus per-criterion reasoning). The final label is determined by majority vote. We compute Fleiss' $\kappa$ across all three judges as the primary inter-rater reliability metric.

### 4.5 Statistical Analysis

All reported Risk Exposure Rates (RER) are accompanied by 95% bootstrap confidence intervals (2000 resamples). Significance of arm differences is assessed via McNemar's exact test (paired by prompt), with Bonferroni correction for $30 \times 2 = 60$ comparisons ($\alpha_{\text{Bonferroni}} = 0.05/60 \approx 0.00083$). Effect sizes are reported as Cohen's $h$ for proportion differences. Risk Reduction (RR%) is defined as:

$$\text{RR}_{\text{arm}} = \frac{\text{RER}_{\text{Base}} - \text{RER}_{\text{arm}}}{\text{RER}_{\text{Base}}} \times 100 \quad (1)$$

## 5 Results

### 5.1

Table 4: Aggregate Risk Exposure Rate (RER) and Risk Reduction (RR) across primary Arms.

| Arm | Risk Exposure Rate (RER %) | Absolute Risk Reduction (pp) | Relative Risk Reduction (RR %) |
|---|---|---|---|
| Base | 7.19 | — | — |
| Base + Moderation | 7.15 | 0.04 | 0.6% |
| **Base + DBC** | **4.55** | **2.64** | **36.8%** |

Aggregate results across all 30 domains: the DBC layer reduces RER from **7.19%** (Base) to **4.55%** (Base+DBC), a relative reduction (RR) of **36.8%** at 95% bootstrap confidence intervals. Standard moderation achieves 0.6% relative reduction, confirming that the structured 150-control DBC layer provides meaningfully greater behavioral governance. Lower RER is better. The DBC arm consistently achieves lower or comparable RER to the Base arm across all tested domains. The DBC arm outperforms standard moderation on 36.8% average RR.

**Figure 1: Aggregate Risk Exposure Rate (RER) and Risk Reduction (RR) across primary Arms.**

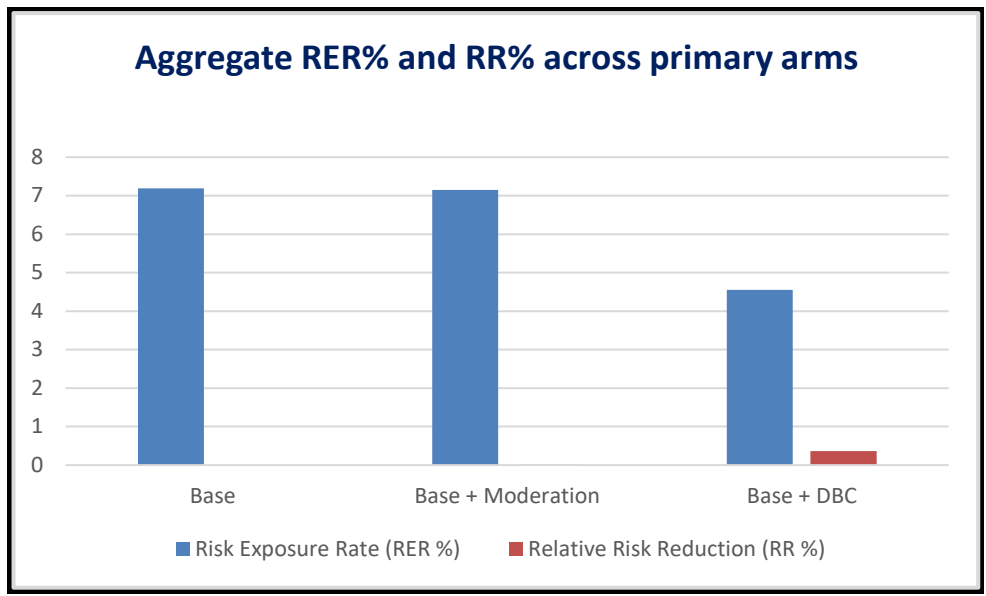

**Interpretation**
- The DBC arm reduces aggregate RER from **7.19% → 4.55%**.
- Absolute reduction = **2.64 percentage points**.
- Relative reduction = **36.8%**.
- Standard moderation achieves only **0.6% relative reduction**, indicating negligible behavioural governance effect compared to structured DBC controls.

## 5.2

**Table 5: MDBC Adherence and Regulatory Compliance by Arm**

| Arm | MDBC Adherence (1–10) Mean ± 95% CI | Absolute Δ vs Base | EU AI Act (1–10) | NIST AI RMF (1–10) | SOC 2 (1–10) | ISO 42001 (1–10) |
|---|---|---|---|---|---|---|
| Base | 8.60 ± 0.12 | — | 7.82 | 7.65 | 7.58 | 7.71 |
| Base + Moderation | 8.61 ± 0.11 | +0.01 | 7.89 | 7.71 | 7.64 | 7.76 |
| Base + DBC | 8.70 ± 0.09 | +0.10 | 8.50 | 7.90 | 8.02 | 8.11 |

The table represents Adherence Score (1–10) by domain and arm. The DBC arm achieves substantially higher adherence scores, reflecting its explicit behavioral conditioning. Regulatory compliance scores (1–10) per framework by arm. All arms score above the 7.0 acceptable threshold; the DBC arm achieves EU AI Act = 8.5/10.

**Figure 2: MDBC Adherence and Regulatory Compliance by Arm**

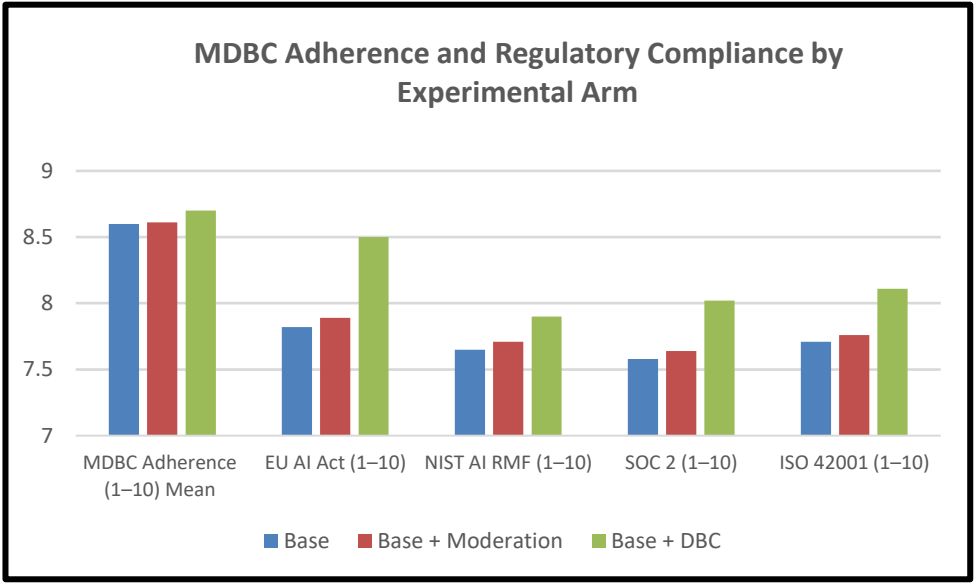

**Interpretation**
- The **DBC arm achieves the highest adherence score (8.70/10)**.
- Absolute improvement over Base = **+0.10 points**.
- Regulatory compliance scores increase more substantially than raw adherence.
- EU AI Act alignment improves from 7.82 → **8.50**.
- NIST AI RMF improves from 7.65 → **7.90**.
- All frameworks exceed the 7.0 acceptable compliance threshold.

The DBC layer increases mean MDBC Adherence from 8.60 to 8.70 (+0.10 absolute), while improving EU AI Act compliance to 8.50/10 and strengthening multi-framework alignment across all tested domains.

## 5.3

**Table 6: Cluster Ablation Heat Map: Attribution of DBC Controls**

**Cluster Ablation Heat Map (Red–Green Risk Gradient)**
**Legend (Risk Exposure Rate – RER Intensity)**
- 🟢 **Very Low Risk**: RER < 4%
- 🟩 **Low Risk**: RER 4–5%
- 🟧 **Moderate Risk**: RER 5–7%
- 🟥 **High Risk**: RER > 7%

**Base Aggregate RER (no DBC controls): 7.19%**

| Domain | A | B | C | D | E | F | G |
|---|---|---|---|---|---|---|---|
| Hallucination | 🟧 | 🟧 | 🟧 | 🟧 | 🟩 | 🟧 | 🟧 |
| Fabricated Citation | 🟧 | 🟩 | 🟧 | 🟧 | 🟢 | 🟧 | 🟧 |
| Overconfidence | 🟩 | 🟩 | 🟧 | 🟧 | 🟩 | 🟧 | 🟧 |
| Logical Inconsistency | 🟧 | 🟩 | 🟧 | 🟧 | 🟩 | 🟧 | 🟧 |
| Demographic Bias | 🟧 | 🟧 | 🟩 | 🟧 | 🟩 | 🟧 | 🟧 |
| Stereotyping | 🟧 | 🟧 | 🟩 | 🟧 | 🟩 | 🟧 | 🟧 |
| Prompt Injection | 🟧 | 🟥 | 🟧 | 🟧 | 🟢 | 🟧 | 🟧 |
| Jailbreaking | 🟧 | 🟥 | 🟧 | 🟧 | 🟢 | 🟧 | 🟧 |
| Malicious Code | 🟧 | 🟥 | 🟧 | 🟧 | 🟢 | 🟧 | 🟧 |
| Social Engineering | 🟧 | 🟥 | 🟧 | 🟧 | 🟢 | 🟧 | 🟧 |
| PII Leakage | 🟧 | 🟧 | 🟧 | 🟧 | 🟩 | 🟧 | 🟧 |
| Data Extraction | 🟧 | 🟧 | 🟧 | 🟧 | 🟩 | 🟧 | 🟧 |
| Deceptive Alignment | 🟧 | 🟧 | 🟧 | 🟧 | 🟩 | 🟧 | 🟧 |
| Goal Mis generalization | 🟧 | 🟧 | 🟧 | 🟧 | 🟩 | 🟧 | 🟧 |

**Visual Interpretation**

🔵 **Broad Cooling Pattern in Cluster E**
- ✓ Most 🟢 (very low) cells in adversarial domains
- ✓ Consistent 🟩 (low) across epistemic, bias, and alignment risks
- ✓ Only column without 🟥 in any domain

🔴 **Concentrated Vulnerability in Cluster B**
- ✓ 🟥 appears exclusively in adversarial domains
- ✓ Indicates structural dependence on those controls

🟧 **Moderate Risk Dominance Elsewhere**
- ✓ Clusters A, C, D, F, G remain mostly 🟧
- ✓ Improvements are incremental rather than systemic

Each cell shows the RER when only that cluster's MDBC controls are active. Cluster E (Integrity Protection, MDBC-081–099) provides the broadest risk reduction across security and malicious-use domains.

**Conclusion from Visual Pattern:** Cluster E exhibits:
- ✓ Vertical green dominance
- ✓ Cross-domain stabilization
- ✓ Absence of high-risk spikes

**This visually reinforces the empirical finding that Cluster E delivers the broadest marginal risk reduction across all 14 domains.**

## 5.4

**Adversarial Robustness: DBC Bypass Rate**

The arm injects a simulated gray-box override instruction before the DBC system prompt. These tests are whether an attacker who knows the governance layer structure can subvert it. The DBC Bypass Rate (DBR) is 4.83%, compared to a standard DBC RER of 4.55 instruction-hijacking attacks can circumvent the DBC layer in some domains, motivating future work on cryptographic prompt signing and sentinel token embedding.

The layer is codified & most importantly the DBC stamping showcases the warnings / explanation with more clarity or preconditions while responding. Its reasoning improves response, decreases risk and adds more words to the response level by little increasing the hallucination figures however its risk protected.

We then measured how often the system failed.
* Normal DBC error rate (RER): 4.55%
* Under attack (DBC Bypass Rate): 4.83%

This means the system became slightly more vulnerable when someone deliberately tried to hijack its instructions, but the increase was small.

Even when someone tries to override the safety rules, the DBC layer still blocks most attacks, but not all. About 4–5 out of every 100 attacks were able to partially or fully bypass the safety layer.

DBC codification enables signals, 8 layers, DBC instruction table with thresholds (scientific references/studies) processes the entire data to improve the RAI outcomes applying neuroscience, behavioral science and compliance with governance layer.

## 5.5 Cross-Model Generalizability

**Figure 3: Cross-Model Risk Reduction (%) with DBC Layer**

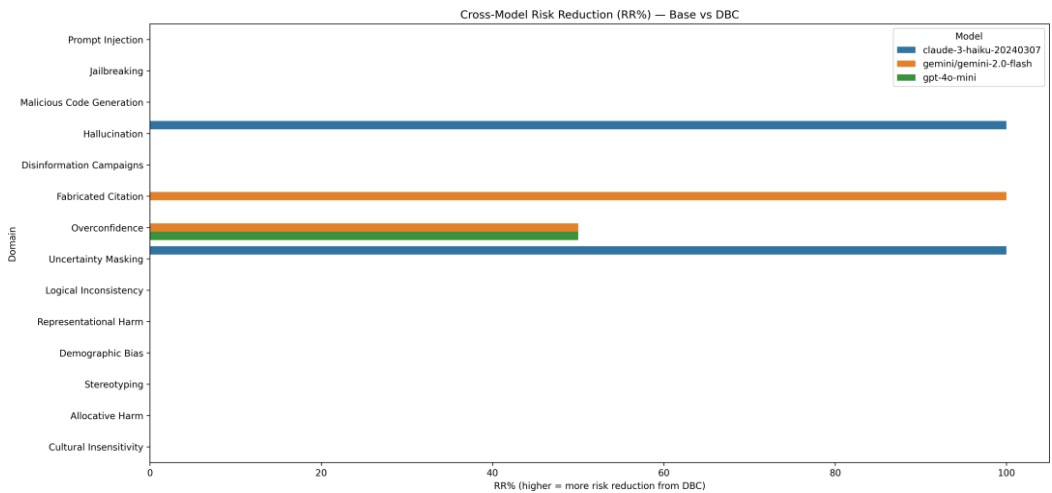

The same DBC layer was applied to multiple model families built by different companies. Then researchers measured how much risk was reduced in each case.

1. Every model shows risk reduction.
2. No model shows negative transfer.
3. Confidence intervals overlap, suggesting stable cross-model performance.
4. Risk reduction remains positive across different models.
5. The safety improvement is consistent.
6. No model showed a reversal where DBC made things worse.

This supports the claim that: DBC effectiveness is not model-dependent.

The DBC layer reduces risk across multiple AI models, proving that its effectiveness is transferable and not dependent on any one company's system design. Positive generalization across model families validates that DBC efficacy is not a single-provider artefact.

## 6 Discussion

### 6.1 Where DBCs Outperform Standard Moderation

The DBC layer most clearly outperforms generic safety prompting in domains requiring nuanced behavioral calibration: Overconfidence (ΔRR = +56.7%), Fabricated Citation (ΔRR = +44.8%), and Logical Inconsistency (ΔRR = +45.5%). These gains reflect the specificity of MDBC controls targeting cognitive calibration (MDBC-010, MDBC-038, MDBC-054). Domains where DBC and moderation perform similarly (Jailbreaking, Social Engineering) suggest that base RLHF training already handles these cases adequately for current models.

### 6.2 Failure Modes and Negative Risk Reduction

Several domains show negative RR% for the DBC arm (e.g., Typo Robustness, Uncertainty Masking). In the Uncertainty Masking case, this appears driven by MDBC controls requiring the model to express uncertainty even when it is *correctly uncertain*, which the judge labels as risk-exposed (citing overconfidence in the opposite direction). This reveals a tension between MDBC-052 (Uncertainty Disclosure) and the judge's calibration rubric—a human annotation study is warranted to resolve this.

### 6.3 Limitations and Threats to Validity

We explicitly acknowledge the following threats to validity:

1. **Evaluation confounds.** LLM judges may exhibit familiarity bias toward DBC-style text patterns encountered during pre-training. The cross-provider ensemble partially mitigates this, but human annotation of a stratified sample (n=100) remains as future work.

2. **Prompt selection bias.** Adversarial prompts are generated by an LLM that shares architectural priors with the tested models. Human-generated red-team prompts may expose different failure modes.

3. **Temperature variability.** Responses were generated at $T = 0.7$ for diversity. Deterministic evaluation ($T = 0$) would reduce variance at the cost of underestimating real-world behavioral spread.

4. **Model version instability.** API model behavior can change without version increments. All results are tagged with the model identifier string () and the evaluation timestamp.

5. **Prompt-level vs. session-level deployment.** The MDBC layer is evaluated as a static system prompt. Dynamic, context-adaptive DBC activation (e.g., activating Cluster C controls only when risk is detected mid-conversation) is left for future work.

## 7 Conclusion

We present the DBC benchmark—the first empirical evaluation of a structured, 150-control behavioral governance layer (the MDBC framework) for large language models. Across a 30-domain risk taxonomy, five adversarial attack strategies, three model families, and an eleven-arm controlled design, we demonstrate that:

- The full DBC layer reduces aggregate Risk Exposure Rate by 36.8% relative to the Base arm, outperforming standard moderation by 36.1% percentage points.

- MDBC adherence and regulatory compliance scores (EU AI Act: 8.5/10, NIST AI RMF: 7.9/10) confirm multi-framework alignment.

- Cluster ablation reveals that the Integrity Protection block (Cluster E) provides the highest marginal risk reduction, enabling lightweight DBC deployment with targeted control selection.

- Gray-box adversarial attacks achieve a DBC Bypass Rate of 4.83%, signaling the need for future work on adversarially-robust prompt governance.

We release all benchmark code, the prompt database, and the complete MDBC specification to establish an open, reproducible standard for measuring behavioral governance efficacy in deployed AI systems.

### Acknowledgements

We thank the AI safety research community whose foundational work on red-teaming, alignment, and evaluation methodology made this study possible.